\documentclass[journal]{IEEEtai}

\usepackage[colorlinks,urlcolor=blue,linkcolor=blue,citecolor=blue]{hyperref}

\usepackage{color,array}
\usepackage{amsmath,amsfonts}
\usepackage{algorithm}
\usepackage{array}
\usepackage[caption=false,font=normalsize,labelfont=sf,textfont=sf]{subfig}
\usepackage{textcomp}
\usepackage{stfloats}
\usepackage{url}
\usepackage{verbatim}
\usepackage{graphicx}
\usepackage{cite}
\usepackage{pifont}
\usepackage[table]{xcolor}
\usepackage{tabu}                     
\usepackage{multirow}                 
\usepackage{multicol}                 
\usepackage{multirow}                
\usepackage{float}                    
\usepackage{makecell}                 
\usepackage{booktabs} 
\usepackage{algorithmicx,algorithm}
\usepackage[noend]{algpseudocode}
\usepackage{algpseudocode}

\newcommand{\eg}{\textit{e}.\textit{g}.}
\newcommand{\etal}{\textit{et al}.}
\newcommand{\ie}{\textit{i}.\textit{e}.}
\newcommand{\etc}{\textit{etc}}

\begin{document}

\title{Test-Time Adaptation for Nighttime Color-Thermal Semantic Segmentation}

\author{Yexin Liu, Weiming Zhang, Guoyang Zhao, Jinjing Zhu, Athanasios V. Vasilakos, and Lin Wang$^\dagger$
\thanks{Manuscript received April 19, 2023. $^\dagger$ corresponding author}
\thanks{Y. Liu, W. Zhang, and Jingjin Zhu are with the Artificial Intelligence Thrust, HKUST(GZ), Guangzhou, China. E-mail:yliu292@connect.hkust-gz.edu.cn, zweiming996@gmail.com, and 	zhujinjing.hkust@gmail.com}
\thanks{G. Zhao is with the Robotics and Autonomous Systems Thrust, HKUST(GZ), Guangzhou, China. E-mail:gzhao492@connect.hkust-gz.edu.cn}
\thanks{Athanasios V. Vasilakos is with the Center for AI Research (CAIR), University of Agder(UiA), Grimstad, Norway. Email: thanos.vasilakos@uia.no}
\thanks{L. Wang is with the Artificial Intelligence Thrust, HKUST(GZ), Guangzhou, and Dept. of Computer Science and Engineering, HKUST, Hong Kong SAR, China. E-mail: linwang@ust.hk 
}}

\markboth{Journal of IEEE Transactions on Artificial Intelligence, Vol. 00, No. 0, Month 2020}
{First A. Author \MakeLowercase{\textit{et al.}}: Bare Demo of IEEEtai.cls for IEEE Journals of IEEE Transactions on Artificial Intelligence}

\maketitle

\begin{abstract}
The ability to scene understanding in adverse visual conditions, \eg, nighttime, has sparked active research for color-thermal semantic segmentation. 
However, it is essentially hampered by two critical problems: 1) the day-night gap of color images is larger than that of thermal images, and 2) the class-wise performance of color images at night is not consistently higher or lower than that of thermal images.
We propose the \textbf{first} test-time adaptation (TTA) framework, dubbed \textbf{Night-TTA}, to address the problems for nighttime color-thermal semantic segmentation \textbf{without access to the source (daytime) data} during adaptation.
~Our method enjoys three key technical parts.
Firstly, as one modality (\eg, color) suffers from a larger domain gap than that of the other (\eg, thermal),
Imaging Heterogeneity Refinement (\textbf{IHR}) employs an interaction branch on the basis of color and thermal branches to prevent cross-modal discrepancy and performance degradation.
Then, Class Aware Refinement (\textbf{CAR}) is introduced to obtain reliable ensemble logits based on pixel-level distribution aggregation of the three branches.
In addition, we also design a specific learning scheme for our TTA framework, which enables the ensemble logits and three student logits to collaboratively learn to improve the quality of predictions during the testing phase of our Night TTA. 
Extensive experiments show that our method achieves state-of-the-art (SoTA) performance with a 13.07\% boost in mIoU.
\end{abstract}


\begin{IEEEImpStatement}
Night-time segmentation is a critical task for autonomous driving under challenging visual conditions. Existing methods mostly focus on daytime segmentation with perfect illumination. This has inspired active research on color-thermal semantic segmentation as thermal cameras are less affected by illumination changes and can complement color modality. 
However, thermal images suffer from a lack of large-scale labeled datasets, which are labor-intensive to obtain. TTA allows for the on-the-fly adaptation to different target domains at the testing phase while protecting data privacy. In light of this, we propose the first TTA framework that achieves SoTA nighttime color-thermal segmentation performance at the testing phase without relying on the source (daytime) data. This is practically valuable for real-world application scenarios. 
The proposed method presents a robust solution for all-day scene understanding, which may hopefully inspire more research in the community. Our project code will be available at \url{https://vlis2022.github.io/nighttta}.
\end{IEEEImpStatement}

\vspace{-10pt}
\section{Introduction}
\vspace{-10pt}
\label{sec:intro}

\begin{figure}[t]
    \captionsetup{font=small}
    \centering
    \includegraphics[width=0.5\textwidth]{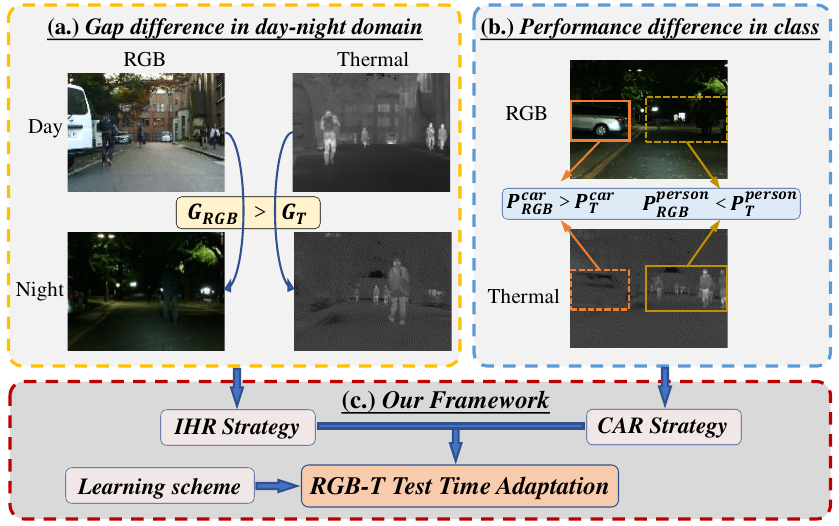}
    \vspace{-19pt}
    \caption{Our Night-TTA addresses two crucial problems for nighttime color-thermal segmentation. 
    \textbf{(a)} demonstrates that the day-night domain gap of each modality ($G_{color}$,$G_{T}$) and the disparity between two domain gaps due to the imaging heterogeneity ($G_{color}$ $>$  $G_{T}$). \textbf{(b)} shows that color and thermal have different advantages in the performance among classes. Hence, the three branches' weights must be reassigned due to class-wise prediction heterogeneity. In \textbf{(c)}, we propose IHR, CAR, and a learning scheme to achieve TTA for nighttime color-thermal segmentation.}
    \label{cover figure}
    \vspace{-15pt}
\end{figure}

\IEEEPARstart{R}{ecent} years have witnessed the success of deep neural networks (DNNs) for color image semantic segmentation, which is crucial for the scene understanding, \eg, autonomous driving~\cite{vertens2020heatnet,chen2017deeplab,li2018pyramid,dou2020unpaired,badrinarayanan2017segnet}. 
However, models trained in favorable lighting conditions show poor generalization ability to the nighttime data. Thus, nighttime image semantic segmentation has become a challenging problem.
Recently, increasing attention has been paid to thermal images because they are inherently robust to illumination changes and may complement semantic information to the color images (especially nighttime images)~\cite{bilodeau2014thermal,li2020segmenting,kim2021ms,kutuk2022semantic,xiong2021mcnet,chen2020nv,wang2019thermal}. For instance, \cite{feng2023cekd, gan2023unsupervised} concentrate on enhancing the effectiveness of thermal image segmentation through knowledge distillation, facilitating the transfer of knowledge from color images to the thermal domain. This has sparked research for supervised ~\cite{sun2019rtfnet,xiong2021mcnet} and unsupervised ~\cite{vertens2020heatnet,kim2021ms} color-thermal semantic segmentation as both modalities can compensate for each other’s deficiencies.

However, existing supervised methods necessitate well-label annotations, particularly for thermal images captured during nighttime, which poses significant labor-intensive challenges.
Meanwhile, most unsupervised methods (\eg, unsupervised domain adaptation (UDA)) entail the drawbacks of time-consuming offline domain adaptation training, and its performance is greatly affected by the domain gap, leading to limited adaptation in diverse testing environments.
Therefore, it is non-trivial as only the nighttime color-thermal data is available under a limited overhead for adaption. This motivates us to explore a suitable adaptation strategy for nighttime color-thermal semantic segmentation. 

Test-Time Adaptation (TTA) ~\cite{prabhu2021s4t,wang2020tent,shin2022mm,zhang2022auxadapt,karani2021test,valanarasu2022fly} presents a practical domain adaptation approach that enables the seamless adaptation of pre-trained models to the target domain in real-time during the testing phase. TTA is different from the UDA-based semantic segmentation setting~\cite{vertens2020heatnet,kim2021ms}: TTA does not need to access source data during adaptation. Moreover, the TTA framework can achieve privacy protection while allowing for on-the-fly adaptation to different target domains during the testing phase without the need for offline domain adaptation training. This is practically valuable for real-world applications. However, directly extending existing TTA methods to color-thermal semantic segmentation leads to less optimal performance, as demonstrated in Tab. \ref{tab:comparative studies_1} in the experiments. For example, entropy minimization of TENT~\cite{wang2020tent} generates overconfident predictions. Therefore, applying it individually to color and thermal branches aggravates the color-thermal discrepancy. 

\textbf{Motivation:} In this paper, we, for the \textit{first} time, explore a TTA framework for nighttime color-thermal semantic segmentation \textit{without access to the source (daytime) color-thermal data}. 
Our work addresses two challenges for nighttime color-thermal semantic segmentation arising from the modality differences during TTA, as shown in Fig.~\ref{cover figure}. 
\textbf{(1)} Due to the different imaging mechanisms, the day-night domain gap, denoted as $G_{color}$, of color images is larger than that, denoted as $G_{T}$, of the thermal images (See Fig.~\ref{cover figure}(a)). This unbalanced difference between $G_{color}$ and $G_{T}$ leads to the considerable cross-modal discrepancy and performance degradation in the adaption process.
We refer to this issue as \textit{imaging heterogeneity}. \textbf{(2)} Existing color-thermal segmentation methods,~\eg,~\cite{vertens2020heatnet,kim2021ms,zhou2021gmnet,lan2022mmnet,xu2021attention}, apply the same weights to all classes. However, we find that the class-wise performance at night (denoted as $P_{color}$) of color images is not consistently higher or lower than that of the thermal images (denoted as $P_{T}$). Therefore, these methods might neglect the discriminative features of the modalities with smaller weights during the color-thermal nighttime segmentation ensemble process. An example is shown in Fig.~\ref{cover figure}(b), where the performance $P_{T}^{person}$  on the class `person' in the thermal image is larger than $P_{color}^{person}$ of the color image. We refer to this as \textit{class-wise prediction heterogeneity}.

To address aforementioned challenges, we propose a novel nighttime TTA framework, called \textbf{Night-TTA}, which consists of three key technical components: \textbf{(1}) Imaging Heterogeneity Refinement (\textbf{IHR}) (Sec.~\ref{sec 3.1}) and \textbf{(2}) Class Aware Refinement (\textbf{CAR}) (Sec.~\ref{Sec_CAR}) and \textbf{(3)} a learning scheme (Sec.~\ref{sec 3.3}), as shown in Fig.~\ref{cover figure}(c). For IHR, we propose an interaction branch to obtain the color-thermal cross-modal invariant feature to prevent the performance degradation in the adaptation process caused by the difference in the cross-modal domain gap ($G_{color}$$>$$G_{T}$).
Specifically, we first take the color-thermal image pairs as input to the interaction branch and then use the two encoders to obtain the color and thermal features that need to be fused. However, directly fusing the color and thermal features induces inconsistent noises due to the private information in the two individual branches. Therefore, we introduce a novel cross-modal shared attention (CMSA) module to aggregate the cross-modal invariant features while suppressing the noisy ones between the two modalities. 

The CAR strategy employs an element-wise entropy-based fusion (EEF) module to generate reliable ensemble logits. This subtly avoids neglecting the discriminative feature information of each class in each branch. Specifically, we first evaluate Shannon entropy in the channel dimension of each student's logits. Then, we re-weight the students' logits to generate more reliable ensemble logits (\ie, teacher) based on the pixel-level distribution of three students. By performing pixel-wise re-weight on the logits of the three branches, the performance advantages of different modalities in different classes can be utilized, and more reliable ensemble logits can be obtained. 

Lastly, we present a novel learning scheme to overcome the potential problematic segmentation results during TTA. By utilizing the reliable ensemble logits generated by the EEF module as a self-supervised signal, we enable three student networks to learn from each other through online distillation~\cite{wang2021knowledge, zhang2018deep, hinton2015distilling} during the adaptation process. This allows our Night-TTA model to fully utilize the discriminative information in each branch, thus preventing the ensemble logits from making false predictions among the categories.

\textbf{Contribution:}  In summary, our major contributions are four-fold: (\textbf{I}) We make the \textit{first} attempt and propose a novel TTA framework for color-thermal semantic segmentation. (\textbf{II}) We propose an IHR strategy with the CMSA module, to reduce the imaging heterogeneity during TTA. We also propose the CAR strategy to take advantage of the segmentation performance of different modalities in different classes and then generate reliable ensemble logits. 
(\textbf{III}) For cross-modal ensemble distillation of our Night-TTA framework, we propose a novel learning scheme to achieve cross-modal ensemble distillation in the testing phase. (\textbf{IV}) Extensive experiments demonstrate that our method significantly surpasses the baselines and prior methods (at least 3.11\% mIoU improvement on the MF-1 dataset, and 2.69\% mIoU  improvement on the KP dataset).


\begin{figure*}[t!]
    \centering
    \includegraphics[width=.92\textwidth]{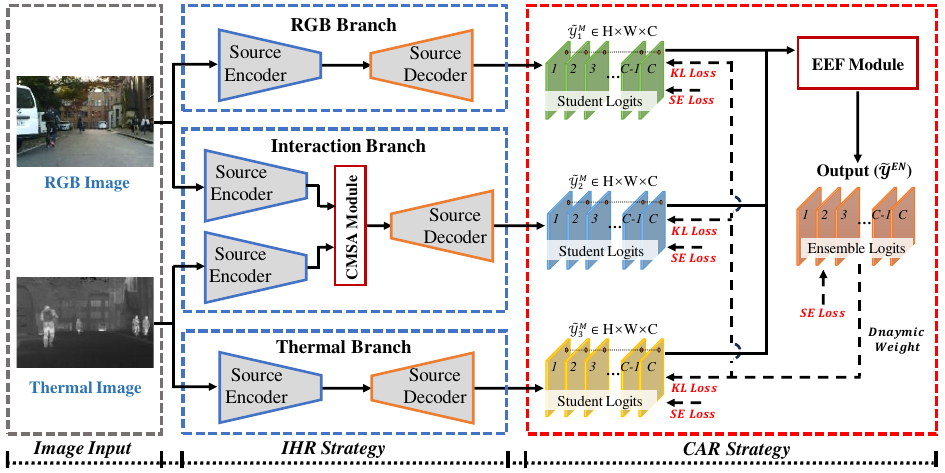}
    \vspace{-10pt}
    \caption{\small{\textbf{Overview of Night-TTA framework}, which consists of Color, Thermal, and Interaction branches (\ie, students). Importantly, we propose Image Heterogeneity Refinement (IHR) and Class Aware Refinement (CAR). The IHR employs an interaction branch with a CMSA module to maintain the shared information to ensure reliable predictions for either color or thermal branch.
    The CAR is buttressed by the EEF module to generate more reliable ensemble logits (\ie, teacher) to learn the discriminative feature of each class in color-thermal modalities. We also propose the training and testing schemes of our TTA framework.}}
    \label{framework}
    \vspace{-10pt}
\end{figure*}

\section{Related work}

\label{sec:Related work}
\noindent \textbf{Color-Thermal Image Semantic Segmentation.}
Recent years have witnessed significant progress in autonomous vehicle scene understanding due to advancements in deep neural networks. Yet, challenges persist in intricate scenarios like nighttime conditions, leading to accuracy reduction and highlighting limitations in these methodologies. To tackle these issues, generative adversarial networks (GANs)\cite{romera2019bridging} and adversarial training strategies\cite{Wu_2021_CVPR} have been proposed to bridge the gap between day and night image domains. GAN-based methods hinge on the effectiveness of image transfer models for semantic segmentation, while adversarial training recalibrates images to align intensity distributions across domains. A rising trend is the attention towards thermal images, valued for their resilience to illumination changes and potential to enhance semantic information from color images. This trend has spurred research in color-thermal semantic segmentation, exploiting modalities to offset inherent limitations.

Color-thermal segmentation methods can be divided into two main categories: supervised methods and unsupervised methods. The former includes the fusion of multi-modalities using multiple encoders with a shared decoder~\cite{ha2017mfnet,sun2019rtfnet,shivakumar2020pst900,sun2020fuseseg,zhou2021gmnet,lan2022mmnet,xu2021attention,wang2020cross,wang2019rgb} and the translation between the RGB and thermal images~\cite{zhang2021abmdrnet}. MFNet~\cite{ha2017mfnet} extracts features from the color and thermal images using two encoders and expands the receptive field by using the 'mini-inception' module. ABMDRNet~\cite{zhang2021abmdrnet} solves the problems of multimodal disparity and multi-scale contextual information fusion by using a bridging-then-fuse strategy to obtain more discriminative cross-modal information.
UDA-based methods, \eg, HeatNet~\cite{vertens2020heatnet}, propose a teacher-student learning method~\cite{wang2021knowledge} to transfer the knowledge from the daytime color image domain to the nighttime thermal image domain to avoid expensive nighttime image annotation. MS-UDA~\cite{kim2021ms} enhances the performance of thermal segmentation by transferring knowledge from color to thermal modality.
By contrast, we propose the first  color-thermal TTA framework that consists of triple student networks for nighttime image semantic segmentation without access to the source domain (daytime) data. Moreover, our TTA framework not only considers the difficulty of the domain gap faced by UDA but also proposes and solves the two novel problems based on the differences between modalities.



\noindent \textbf{Test-Time Adaptation (TTA).}
TTA methods enable the model to adapt quickly to the target domain, which does not require access to source domain data.\cite{chi2021test,sun2020test}. TTA has been applied to unimodal\cite{wang2020tent,prabhu2021s4t,zhang2022auxadapt} and cross-modal\cite{shin2022mm,jaritz2020xmuda} segmentation tasks.
For the former task, the typical model Tent\cite{wang2020tent} presents an entropy minimization strategy to optimize afﬁne parameters during testing.
For the Cross-modal segmentation task, xMUDA\cite{jaritz2020xmuda} allows the 2D and 3D modalities to learn from each other via imitation, disentangled from the segmentation objective to prevent false predictions. MM-TTA\cite{shin2022mm} proposes two complementary modules to obtain and select more reliable pseudo-labels (from 2D and 3D modalities) as self-learning signals during TTA. 
However, directly using previous TTA methods for color-thermal semantic segmentation leads to less optimal performance. Therefore, we propose the IHR and CAR strategies to make our color-thermal TTA framework more robust and generalized, with a unique learning scheme that can perform better in both the training and testing phases.


\noindent \textbf{Ensemble distillation.}
Compared with the standard knowledge distillation (KD) paradigm\cite{hinton2015distilling,gou2021knowledge,alkhulaifi2021knowledge}, online KD (ensemble distillation)\cite{zhang2018deep,anil2018large,zhu2018knowledge,song2018collaborative,guo2020online,wu2021peer} enables efficient and single-stage training via collaborative learning among the student networks. Collaborative learning relies on two main ways:  students learn from each other \cite{zhang2018deep,anil2018large,song2018collaborative} or generate ensemble logits to supervise their learning\cite{zhu2018knowledge,guo2020online,wu2021peer}.
The former methods facilitate peers' mutual learning by sharing knowledge among the student networks. For example, CLNN\cite{song2018collaborative} allows multiple classifier heads to share intermediate-level representation for collaborative learning to reduce generalization errors.
The latter methods focus on generating ensemble logits that update each student's network based on the contributions shared by the students. In particular, \cite{guo2020online,wang2021knowledge} select the logits based on the cross-entropy loss of each student with the true label. However, we cannot access the labels during test time. Therefore, we propose the CAR strategy to generate reliable ensemble logits, which considers the different class-wise performance between the two modalities.

\section{Method}
\noindent \textbf{Overview.} In multi-modal TTA for color-thermal image semantic segmentation, we consider a source domain dataset, where each sample consists of daytime paired color images ($x_s^{color} \in \mathbb{R}^{H \times W \times 3}$), thermal images ($x_s^{T} \in \mathbb{R}^{H \times W \times 1}$), and corresponding segmentation ground truth (GT). A source model is trained on the labeled source domain dataset. Usually, the source model consists of a color encoder $E_{color}$, a thermal encoder $E_T$, and the decoder $D$ utilized to generate pixel-level semantic labels. The source model can be denoted as $f_\theta=D(E_{color}(x_s^{color}), E_T(x_s^{T}))$. 

Typically, the performance of the source model $f_\theta$ is unsatisfactory when confronted with new test data characterized by a different distribution from the source samples. The primary objective of TTA is to enhance the prediction performance in the target domain by conducting model adaptation solely on unlabeled target data. Specifically, given a target dataset $t$, which comprises nighttime paired color images ($x_t^{color}$) and thermal images ($x_t^{T}$).
The model is updated using $\operatorname*{min}_{\tilde{\theta}}\mathcal{L}(\mathbf{x};\theta),\mathbf{x}\sim t$
, where $\tilde{\theta}\subseteq \theta$ represent the model parameters that should be updated (\eg, batch normalization layer), $\mathcal{L}$ denotes self-supervised loss functions.

Prior research works on TTA have employed the entropy minimization for single-modality (\eg, color image) semantic segmentation~\cite{wang2020tent} or utilized consistency loss and pseudo-labels for cross-modal (\eg, 2D-3D) segmentation~\cite{jaritz2020xmuda,shin2022mm}. 
However, as discussed above, applying existing TTA methods directly to color-thermal semantic segmentation poses challenges due to two main factors: \textit{imaging heterogeneity} and \textit{class-wise prediction heterogeneity}.

To this end, we propose a novel TTA framework for nighttime color-thermal image semantic segmentation. Specifically, as depicted in Fig.~\ref{framework}, the proposed TTA framework consists of color, thermal, and interaction branches, representing three separate student networks.
color, thermal, and interaction branches take the $x_t^{color}$, $x_t^{T}$, and both as the input, respectively.
\textit{There are two novel technical components}: IHR (Sec.~\ref{sec 3.1}) and CAR (Sec.~\ref{Sec_CAR}).
To solve the problems caused by imaging heterogeneity, the IHR employs an interaction branch with a novel cross-modal shared attention (CMSA) module to generate reliable pseudo labels. The CMSA module is introduced before the decoder to aggregate the complementary features and suppress the noisy features of the color and thermal modalities.
To solve the problems caused by class-wise prediction heterogeneity, the CAR is buttressed by an element-wise entropy-based fusion (EEF) module to generate the ensemble logits by aggregating the reliable logits from three branches. We also propose a specific learning scheme that enables the three student networks to collaboratively learn to improve the quality of predictions during adaptation.

\vspace{-10pt}
\subsection{Imaging Heterogeneity Refinement (IHR)}
\begin{figure}[t]
    \centering
    \includegraphics[width=0.5\textwidth]{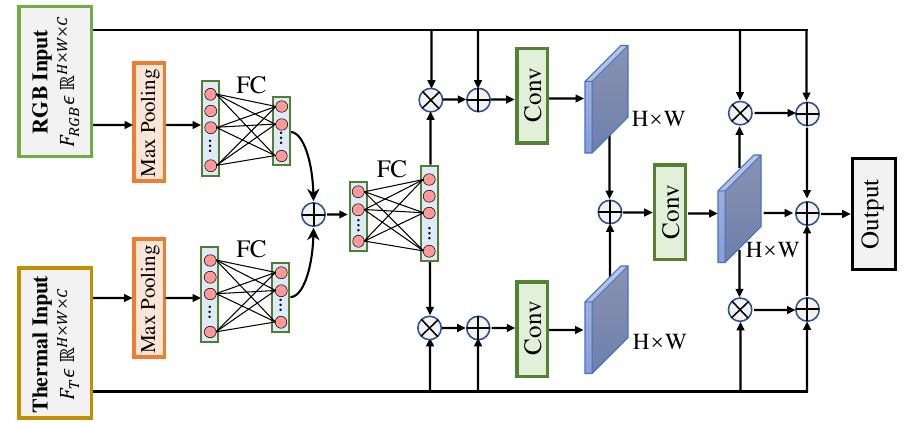}
    \vspace{-18pt}
    \caption{\small{Visualization of the CMSA is depicted. The process involves passing the RGB and thermal inputs through dedicated Maxpooling and Fully-Connected (FC) layers individually. Subsequently, an element-wise 'add' operation, followed by another FC layer, is employed to extract shared-channel features. These extracted features are then subjected to multiplication and addition with the original input to rectify any noisy features. A similar architectural arrangement is utilized to extract shared-spatial features and subsequently rectify any associated noise.}}
    \label{CMSA}
    \vspace{-20pt}
\end{figure}

\label{sec 3.1}

The straightforward fusion of the color and thermal branches leads to a noticeable degradation in the segmentation performance due to the significant domain gap between the two modalities, as evidenced by the results presented in Tab. \ref{tab:ablation studies}. To address this challenge, we propose the integration of an interaction branch to facilitate the extraction of cross-modal invariant features, which are crucial for generating reliable pseudo labels.

Specifically, color images provide abundant textual information that is valuable for segmentation tasks, particularly in well-illuminated daytime scenarios. However, their performance suffers greatly when confronted with adverse lighting conditions. On the contrary, thermal images exhibit robustness to illumination changes but exhibit limitations such as lower resolution and ambiguous object boundaries. Therefore, a direct fusion of color and thermal features may introduce inconsistencies caused by the individual characteristics of each modality, undermining segmentation accuracy.

To mitigate these issues, the introduction of the interaction branch aims to exploit the complementary nature of color and thermal modalities. This branch facilitates the extraction of cross-modal invariant features that are resilient to domain gaps, enabling the generation of more reliable pseudo labels. By integrating these cross-modal invariant features with the individual modalities, we can effectively capture both shared and unique information, leading to improved segmentation performance in color-thermal images.
This may cause generating unreliable pseudo labels. 
For this reason, we design the CMSA module (see Fig.~\ref{CMSA}) to rectify the noisy features and extract the cross-modal invariant features.

For the CMSA, we first embed both color ($F_{color} \in \mathbb{R}^{H \times W \times C}$) and thermal ($F_{T} \in \mathbb{R}^{H \times W \times C}$) features into two individual channel (C) attention vectors ($V_{color}^C\in \mathbb{R}^C$) and ($V_{T}^C\in \mathbb{R}^C$). Unlike~\cite{liu2022cmx}, rectifying features by utilizing the individual vectors, we generate the shared channel attention vectors ( $V_{shared}^C\in \mathbb{R}^C$) by aggregating the vectors from the color-thermal features to maintain the shared features while suppressing the noisy features.
The channel-wise feature rectiﬁcation can be described as:

\begin{equation}
\footnotesize
\begin{aligned}
    F^C_{color}&=V_{shared}^C \odot F_{color} +F_{color}, \\
    F^C_{T}&=V_{shared}^C \odot F_{T} + F_{T}.
  \label{eq:W}
\end{aligned}
\end{equation}

Similar to the channel-wise rectiﬁcation, a shared spatial (S) attention vector ($V_{shared}^S\in \mathbb{R}^{H \times W}$) is embedded to calibrate the local information, which is formulated as follows:
\begin{equation}
\footnotesize
\begin{aligned}
     F^S_{color}&=V_{shared}^S \odot F^C_{color} + F^C_{color},\\
    F^S_{T}&=V_{shared}^S \odot F^C_{T} + F^C_{T}.
  \label{eq:W}   
\end{aligned}
\end{equation}

$F^S_{color}$ and $F^S_{T}$ are the rectiﬁed features after the CMSA module, which will be aggregated to the decoder of the interaction branch. Once obtained the logits in each branch, pseudo-labels are provided for the CAR.

\begin{figure}[t!]
    \centering
    \includegraphics[width=0.5\textwidth]{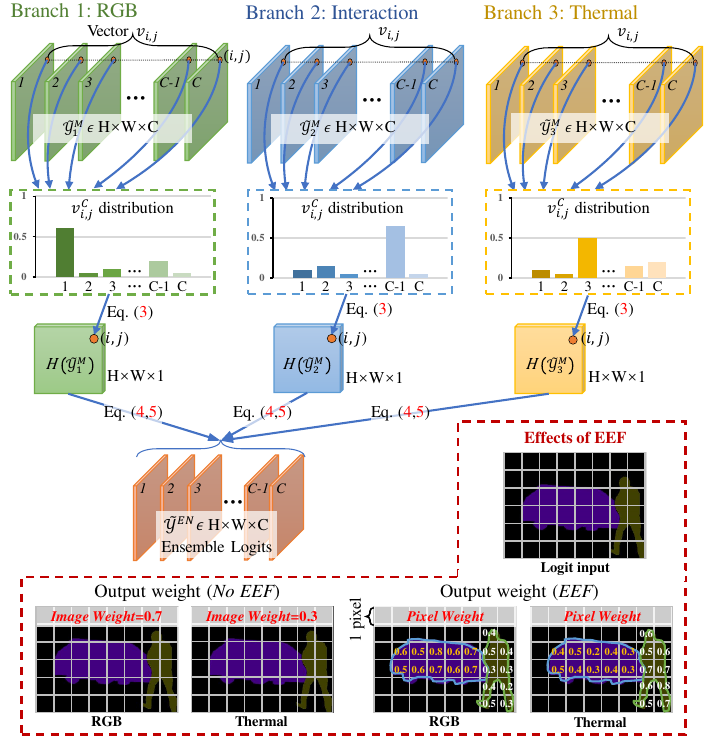}
    \vspace{-15pt}
    \caption{\small{Visualization of the EEF module is presented. The upper section delineates the EEF module's calculation process, while its functionality is elaborated below. Within the EEF module, the procedure involves obtaining the distribution of logits for each branch at every position. Subsequently, element-level entropy is computed. Ultimately, considering the entropy values of the branchless logits, the fusion and screening process yields the logits for each position.}}
    \label{EEF_fig}
    \vspace{-15pt}
\end{figure}

\vspace{-10pt}
\subsection{Class Aware Refinement (CAR)}
\label{Sec_CAR}
To generate ensemble logits,  previous method,~\eg,~\cite{shin2022mm} usually assigns an image-level weight to each branch by measuring the consistency between the cross-modal branches. This may encounter class performance imbalance problems for color-thermal segmentation due to the class-wise prediction heterogeneity in cross-modalities. Take the cross-modal branches as an example (See Fig. \ref{EEF_fig}). We assume that the weights calculated by the existing method for the color and thermal branch are 0.7 and 0.3, respectively. When generating the ensemble logits, all classes in the color branch are assigned a weight of 0.7, while those of the thermal branch are assigned 0.3. This leads to poor segmentation performance for some classes that were originally better segmented in the thermal branch (\ie, person). To alleviate this problem, we propose the EEF module to refine the ensemble logits, as shown in Fig.~\ref{EEF_fig}. 



\subsubsection{Element-wise Entropy-Based Fusion (EEF)}
\label{sec 3.2}
The EEF module uses the outputs of three branches as the input, which are denoted as ${\tilde{y}}_1^M$, ${\tilde{y}}_2^M$, and ${\tilde{y}}_3^M$ (${\tilde{y}}_1^M$, ${\tilde{y}}_2^M$, ${\tilde{y}}_3^M \in \mathbb{R}^{H \times W \times C}$) respectively, where $M \in \left\{s, t\right\} $and C denotes the number of channels. To assign the weight $W_i$  for branch $i$, specifically, the $softmax$ is firstly computed along the channel dimension. Then, we calculate the Shannon entropy ($H({\tilde{y}}_i^M) \in \mathbb{R}^{H \times W \times 1}$) of the logits ${\tilde{y}}_i^M$. For each pixel $(i, j) \in H \times W $, we can obtain a vector $v_{i,j} \in 1 \times C$ consisting of the elements of logits at position $(i, j)$ for all channels. Then, we calculate the Shannon entropy $H(v_{i,j}^C)$ of the vector $v_{i,j}$:

\begin{small} 
\vspace{-6pt}
\begin{equation}
H(v_{i,j}^C)=\sum\limits_{C=1}^{N}softmax(v_{i,j}^C) \cdot log softmax(v_{i,j}^C),
  \label{eq:H}
\end{equation}
\vspace{-6pt}
\end{small}

where $v_{i,j}^C$ denotes the value of vector $v_{i,j}$ in channel C. $H({\tilde{y}}_i^M)$ is composed of the Shannon entropy ($SE$) of all vectors $v_{i,j}$. Assume that the true label at position ($i$, $j$) is $k$. When the value on the $k$-th channel becomes larger, the value on other channels diminishes. Then the cross entropy (CE) loss with the label decreases, which means the segmentation performance becomes better. The ideal probability distribution is that the prediction on the $k$-th channel is close to 1, while the prediction on the other channels is close to 0. In this situation, Shannon entropy will be kept to a relatively small extent. An effective way to generate teacher logits is to re-weight the student's logits based on the element-wise Shannon entropy. For each element in the teacher's logits, the smaller the Shannon entropy in the channel dimension, the greater the weight of the branch. We deﬁne the teacher's logits as the combination of all students' weighted logits. The pixel-wise weights $W_i$ of branch $i$ are calculated as:
\begin{small} 
\begin{equation}
  W_i=\frac{e^{(1-H({\tilde{y}}_i^M))/temp}}{\sum\limits_{i=1}^3 e^{(1-H({\tilde{y}}_i^M))/temp}},
  \label{eq:W}
\end{equation}
\end{small} 
where $W_i \in \mathbb{R}^{H \times W \times 1}$, $temp$ denotes the temperature. Finally, the teacher's logits are as follows:   
\begin{small} 
\begin{equation}
{\tilde{y}}^{EN}=\sum\limits_{i=1}^3 {W_i*\tilde{y}}_i^M.
  \label{eq:H}
\end{equation}
\end{small} 



\vspace{-20pt}
\subsection{Learning Scheme}
\label{sec 3.3}

For TTA, we denote  the updated parameters of the Batch normalization layer of color, interaction, and thermal branch as $\gamma^{color}, \gamma^{Int}, \text{and}~\gamma^{T}$, respectively. Given paired color-thermal images, there are $i$ classes in the image. The predictions of different branches can be denoted as $P_{color}=\{P_{color}^1,{P_{color}}^2,..., {P_{color}}^i\}$, $P_{Int}=\{P_{Int}^1,{P_{Int}}^2,..., {P_{Int}}^i\}$, and $P_{T}=\{P_{T}^1,{P_{T}}^2,..., {P_{T}}^i\}$. During TTA, the class-wise segmentation performance of one branch is not consistently higher or lower than the other branches. For some classes, one branch can achieve the best segmentation performance while the other branch could achieve the best performance in other classes.  Without loss of generality, we consider the case of three classes where the color, interaction, and thermal branch achieves the best performance on class 1, 2, and 3, respectively. The ensemble logits of traditional methods are calculated by $P_{EN}=\frac{P_{color}+P{Int}+P_T}{3}$. Then, the consistency loss $\mathcal{L}_{KL}^{tta}$, which achieves knowledge distillation from ensemble logits to student logits, is used to train the three branches. During TTA, the parameters of the batch normalization layer $\gamma$ are updated by:

\begin{small} 
\begin{equation}
\begin{gathered}
\gamma_t^{color}=\gamma_{t-1}^{color}-\beta \cdot \bigtriangledown_{\gamma}\mathcal{L}_{KL}^{tta}(color,EN)\\
\gamma_t^{Int}=\gamma_{t-1}^{Int}-\beta \cdot \bigtriangledown_{\gamma}\mathcal{L}_{KL}^{tta}(Int,EN)\\
\gamma_t^{T}=\gamma_{t-1}^{T}-\beta \cdot \bigtriangledown_{\gamma}\mathcal{L}_{KL}^{tta}(T,EN)
\label{eq:TTA}
\end{gathered}
\end{equation}
\end{small} 

Based on our assumptions, for class $1$, the entropy of the color branch is smaller than the ensemble logits ($SE(P_{color}) \textless SE(P_{EN})$), whereas the entropy of the interaction and thermal branches are larger than the ensemble logits ($SE(P_{Int}) \textgreater SE(P_{EN})$ and $SE(P_{T}) \textgreater SE(P_{EN})$).  Therefore, although the interaction and thermal branches will improve the segmentation performance, the color branch will have performance degradation after optimization. The other two classes have similar results.

To mitigate the issues mentioned above, we propose the EEF module and a learning scheme (See Fig. \ref{framework}).

During TTA, we consider the teacher logits as the self-training signals to update the model. We define KL loss as $\mathcal{L}_{KL}^{tta}(i,EN)= KL({\tilde{y}}_i^s,{\tilde{y}}^{EN})$ to ensure collaborative learning of these three students. Moreover, to boost the performance of all three student networks, we introduce the Shannon entropy loss $\mathcal{L}_i^{tta}= SE({\tilde{y}}_i^t)$, and $\mathcal{L}_{EN}^{tta}= SE({\tilde{y}}^{EN})$.
For each student network $i$, the final learning objective is:
\begin{small} 
\begin{equation}
\mathcal{L}^{tta}={\sum\limits_{i=1}^3 {\mathcal{L}_i^{tta}}}+\lambda_1  \mathcal{L}_{EN}^{tta}+\lambda_2 {\sum\limits_{i=1}^3 {\mathcal{L}_{KL}^{tta}(i,EN)}}
  \label{eq:loss_test_all},
\end{equation}
\end{small} 
where $\lambda_1$ and $\lambda_2$ are hyperparameters.

\textbf{Dynamic Weighting Each branch.} Existing methods for multi-modal test time adaptation typically assign the same weights to all branches. However, for color-thermal segmentation, the day-night domain gap in color images is more significant than in thermal images. Consequently, utilizing identical weights for all branches can lead to instability during adaptation. To address this issue, we propose a dynamic weighting scheme for these branches, which exclusively affects the loss function without incurring additional computational overhead for model adaptation. Specifically, we introduce weights $\omega_i$ for each branch according to the adaptation extent. 

Measuring the extent of adaptation typically relies on labeled samples, which presents a challenge in our problem scenario where training data is unavailable, and the test samples remain unlabeled. Consequently, quantifying the extent of adaptation becomes non-trivial. To address this issue, we propose a novel approach that leverages ensemble logits to estimate the extent of adaptation. In particular, we initially compute the distance between the student logits and the ensemble logits of each branch within a batch. This computation can be formulated as follows:

\begin{small} 
\begin{equation}
D_i=\displaystyle \frac{1}{B}\sum_{b=1}^B \frac{1}{2}({\mathrm{KL}}({\tilde{y}}^{EN}||{\tilde{y}}^{M})+{\mathrm{KL}}({\tilde{y}}^{M}||{\tilde{y}}^{EN})),
  \label{eq:d}
\end{equation}
\end{small} 

Then we calculate the weights of each branch as follows:

\begin{small}
\begin{equation}
\omega_i = \frac{D_{i}}{\min\{D_{1}, D_{2},D_{3}\}}
  \label{eq:omega}
\end{equation}
\end{small} 

Then, the final objective is :

\begin{small}
\begin{equation}
\mathcal{L}^{tta}={\sum\limits_{i=1}^3 {\omega_i \mathcal{L}_i^{tta}}}+\lambda_1  \mathcal{L}_{EN}^{tta}+\lambda_2 {\sum\limits_{i=1}^3 {\omega_i \mathcal{L}_{KL}^{tta}(i,EN)}}
  \label{eq:loss_final_test_all},
\end{equation}
\end{small} 
where $\lambda_1$ and $\lambda_2$ are hyperparameters. With the EEF module, we can generate ensemble logits with small entropy at the pixel level. Then, for each class $i$, we have $SE(P_{EN}) \textless SE(P_{color})$, $SE(P_{EN}) \textless SE(P_{Int})$, and $SE(P_{EN}) \textless SE(P_{T})$, which means that we have better ensemble logit to train the three branches. Adaptation with our learning scheme can continuously improve the segmentation performance of the three student branches through ensemble distillation, so as to gradually carry out more accurate segmentation results.

\section{Experiments} 

\subsection{Datasets}
{\bf MF dataset.} It contains 1569 images (784 for training, 392 for validation, and 393 for test) in which 820 daytime and 749 nighttime images are mixed in training, validation, and test sets. The resolution of images is 480$\times$640 with annotated semantic labels for 8 classes. 
To evaluate our method, we just drop out the nighttime color-thermal image pairs in the original training and validation sets and drop out the daytime color-thermal image pairs in the original test sets to form a new dataset (410 for training, 205 for validation, and 188 for test), which is denoted as \textbf{MF-1}. For UDA methods, under our investigation, there only exist two UDA methods (HeatNet and MS-UDA) for nighttime image semantic segmentation leveraging color and thermal images. Thus, we compare the segmentation performance with these two methods. For a fair comparison, we use the same training and testing set with MS-UDA: We reorganize the daytime and nighttime images in the MF dataset as training and testing sets (820 daytime images for training and 749 nighttime images for testing ), which is denoted as \textbf{MF-2}. Three categories of labels overlapping the KP dataset (\eg, car, person, and bike) are used for evaluation.

{\bf The modified KP dataset.} The KAIST Multispectral Pedestrian Detection (KP) dataset ~\cite{hwang2015multispectral} is a color-thermal paired urban driving dataset without semantic segmentation labels. Kim \etal\cite{kim2021ms} create a modified KP dataset with manually annotated 503 daytime and 447 nighttime color-thermal image pairs and the pixel-level labels of 19 classes consistent with Cityscapes ~\cite{cordts2016cityscapes}. The resolution of color-thermal image pairs is 512 $\times$ 640 $\times$ 3 and 512 $\times$ 640 $\times$ 1, respectively.

\begin{table}[t!]
\renewcommand\arraystretch{1.5}
\caption{Quantitative comparisons with source-only supervised methods and TTA combination methods based on MF-1 dataset. Night-TTA$^\dagger$ refers to our trained model (source-only).}
\centering
\setlength{\tabcolsep}{1.5mm}
\footnotesize
\begin{tabular}{@{}clccccc@{}}
\hline
Dataset & Method  & Adapt & Car & Person & Bike & mIoU\\
\hline
\multirow{7}{*}{MF-1} & Source-only & - & 70.27 & 42.01 & 8.00 & 40.09\\
& CMNeXt~\cite{zhang2023delivering} & -& 65.89 & 52.06 & 35.41 & 51.12\\
\cline{2-7}

& LAME~\cite{boudiaf2022parameter} & TTA & 41.73 & 32.19 & 13.87 & 33.21\\
& BN-Adapt~\cite{schneider2020improving} &TTA &43.17 & 35.29 & 14.11 & 34.13\\
& Tent\cite{wang2020tent} & TTA & 44.52 & 34.53 & 15.73 & 35.82\\
& xMUDA-pl\cite{jaritz2020xmuda} & MMTTA& 71.09 & 55.81 & 22.71 & 49.87\\
& MMTTA\cite{shin2022mm}& MMTTA & 71.96 & 54.78 & 23.42 & 50.05\\
& \cellcolor{gray!20}\textbf{Night-TTA (Ours)} &\cellcolor{gray!20} MMTTA & \cellcolor{gray!20}\textbf{76.26} &\cellcolor{gray!20}\textbf{ 58.31} & \cellcolor{gray!20}\textbf{24.76} & \cellcolor{gray!20}\textbf{53.16}\\
\bottomrule
\end{tabular}

\label{tab:comparative studies_1}
\end{table}

\begin{table}[t!]
\renewcommand\arraystretch{1.5}
\caption{Quantitative comparisons with UDA methods and TTA combination methods based on MF-2 dataset.}
\centering
\setlength{\tabcolsep}{1.5mm}
\footnotesize
\begin{tabular}{@{}clccccc@{}}
\hline
Dataset & Method  & Adapt & Car & Person & Bike & mIoU\\
\hline
\multirow{3}{*}{MF-2} & HeatNet\cite{vertens2020heatnet} & \multirow{2}{*}{UDA}& 56.4 & 68.8 & 33.9 & 53.0 \\
& MS-UDA\cite{kim2021ms} & & 69.02 & 52.85 & 49.95 & 57.27\\
\cline{2-7}
& \cellcolor{gray!20}\textbf{Night-TTA (Ours)} & \cellcolor{gray!20}TTA & \cellcolor{gray!20}\textbf{74.19} & \cellcolor{gray!20}\textbf{59.63} & \cellcolor{gray!20}\textbf{55.81} & \cellcolor{gray!20}\textbf{63.32}\\
\bottomrule
\end{tabular}
\vspace{-8pt}

\label{tab:comparative studies_2}
\end{table}

\vspace{-10pt}
\subsection{Implementation Details}

\begin{table*}[t]
\renewcommand\arraystretch{1.5}
\caption{Qualitative results on the modified KP dataset.}
\vspace{-8pt}
\centering
\resizebox{\textwidth}{!}{
\begin{tabular}{cccccccccccccccccccccc}
\hline
\rule{0pt}{50pt} 
Method             & Adapt  & \makebox[0.05\textwidth][c]{\rotatebox{90}{road}}  & \rotatebox{90}{sidewalk} & \rotatebox{90}{building} & \rotatebox{90}{wall}  & \rotatebox{90}{fence} & \rotatebox{90}{pole}  & \rotatebox{90}{traffic light} & \rotatebox{90}{traffic sign} & \rotatebox{90}{vegetation} & \rotatebox{90}{terrain} & \rotatebox{90}{sky}   & \rotatebox{90}{Person} & \rotatebox{90}{rider} & 
\rotatebox{90}{car}   & \rotatebox{90}{truck} & 
\rotatebox{90}{bus}   & \rotatebox{90}{train} & \rotatebox{90}{motorcycle} & \rotatebox{90}{bicycle}& mIoU \\ 
\hline
Source & - & 90.16 & 40.31 & 79.03 & 48.99 & 42.9 & 50.46 & 0.15 & 44.52 & 69.41 & 42.3 & 9.56 & 63.58 & 0 & 76.59 & 0.23 & 9.08 & 0 & 5.73 & 17.59 & 36.35\\
\hline
LAME~\cite{boudiaf2022parameter} & TTA & 82.39 & 41.47 & 49.65 & 36.28 & 14.95 & 38.42 & 0 & 16.49 & 60.19 & 28.46 & 0 & 56.28 & 0 & 68.51 & 0 & 29.31 & 0 & 0 & 16.35 & 28.36\\
BN-Adapt~\cite{schneider2020improving} &TTA & 86.26 & 45.89 & 73.07 & 40.56 & 18.36 & 42.77 & 0 & 20.14 & 64.38 & 32.55 & 0.86 & 60.14 & 0 & 72.67 & 0 & 33.91 & 0 & 0 & 20.85 & 32.23\\
Tent\cite{wang2020tent} & TTA & 88.31 & 47.15 & 76.41 & 42.11 & 20.01 & 44.23 & 0 & 22.18 & 66.25 & 34.19 & 1.08 & 62.35 & 0 & 74.92 & 0 & 35.82 & 0 & 0 & 22.37 & 33.55\\
MMTTA\cite{shin2022mm} & MMTTA & 91.04 & 57.38 & 78.9 & 52.61 & 43.9 & 56.77 & 9.65 & 55.62 & 70.56 & 48.81 & 22.93 & 66.22 & 0 & 78.93 & 0.16 & 61.42 & 0 & 28.03 & 38.92 & 45.36 \\
xMUDA-pl\cite{jaritz2020xmuda} & MMTTA & 91.03 & 57.33 & 80.34 & 53.93 & 44.49 & 53.59 & 0.78 & 49.64 & 71.47 & 49.0 & 19.96 & 67.44 & 0 & 78.88 & 0 & 67.95 & 0 & 26.71 & 43.95 & 45.08 \\
\rowcolor{gray!20}
\textbf{Ours} &  MMTTA & 92.41 & 59.78 & 81.52 & 55.93 & 45.27 & 59.75 & 9.47 & 57.19 & 72.37 & 50.36 & 26.79 & 69.53 & 0 & 80.62 & 0 & 71.92 & 0 & 31.58 & 43.22 & 47.77\\
\hline
\end{tabular}}
\label{KP Test set}
\end{table*}

\begin{figure*}[t!]
    \centering
    \includegraphics[width=0.96\textwidth]{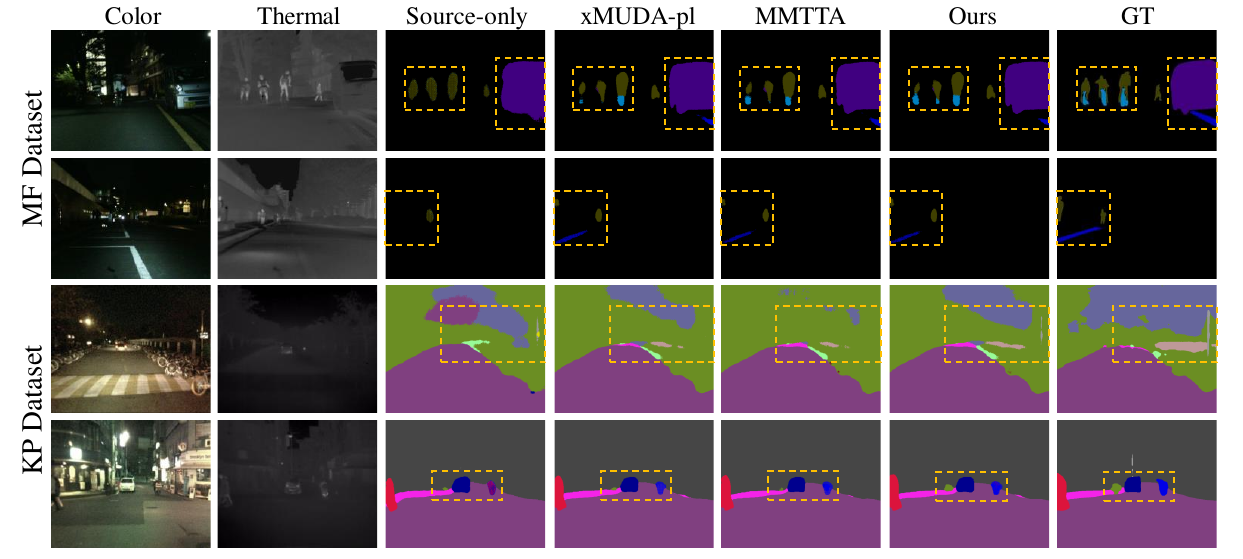}
    \vspace{-10pt}
    \caption{\small{Visual outcomes of our approach are compared with those of existing state-of-the-art methods on both the MF-1 and KP datasets. Notably, the region bounded by the yellow dotted lines highlights areas with significant deviations from the ground truth.}}
    \vspace{-10pt}
    \label{MF-1 Result}
\end{figure*}

The proposed method is implemented using PyTorch libraries with a single A6000 GPU.

{\bf Source model.}
As the first TTA framework for nighttime color-thermal semantic segmentation, our approach adopts a three-branch network structure. Each branch utilizes an untrained encoder and decoder from FEANet \cite{deng2021feanet} (which after the training step already reaches good performance based on a supervised manner) to obtain the logits. We utilize the encoder and decoder from FEANet as the \textit{source} model without changing the network architecture.

{\bf Pre-training the source model.}
In our experiment setting, we want to use daytime data for training and nighttime data for testing. However, the source model from FEANet was trained and tested on day-night mixed dataset which is a different dataset splitting scheme from ours. Therefore, we pre-train the source encoder and source decoder with the source domain dataset.   
For a fair comparison, We follow the training details of FEANet apart from using the original dataset.

{\bf Test-time Adaptation Details.}
We apply the source model that only uses daytime data as training to each branch and use unlabeled nighttime paired data as input for test time adaptation. Similar to previous TTA methods~\cite{wang2020tent,shin2022mm}, we only optimize the batch norm affine parameters for one epoch. The learning rate for three sub-networks is set to $1e^{-5}$. The temperature is set to 2. 
\vspace{-10pt}

\subsection{Comparative Studies}

We evaluate the proposed framework against state-of-the-art TTA methods on the MF-1, MF-2, and modified KP datasets. 


{\bf MF-1 dataset.} We compare our TTA framework with uni-modal and multi-modal TTA frameworks on MF-1 dataset. The quantitative and qualitative results are shown in Tab. \ref{tab:comparative studies_1} and Fig. \ref{MF-1 Result}. The proposed Night-TTA could bring a significant adaptation effect on nighttime color-thermal image semantic segmentation compared to the source model (increases \textbf{13.07} \% mIoU). Specifically, in Tab. \ref{tab:comparative studies_1}, we conduct a comparison of the segmentation performance among different TTA frameworks across three categories: Car, Person, and Bike. Based on the analysis of the experimental data, our TTA framework exhibits a notable improvement in the segmentation performance for all three categories. Moreover, our Night-TTA achieves a substantial performance advantage over both uni-modal TTA methods, with an improvement of over 17.34\% in mIoU. It should be noted that directly applying the uni-model TTA methods would degrade the segmentation performance. Our method also surpasses multi-modal TTA methods with an improvement of over 3.01\% in mIoU.

{\bf MF-2 dataset.} We also compare our method with existing UDA methods. The results are shown in Tab. \ref{tab:comparative studies_2}. In the MF-2 dataset setting, where training is conducted on daytime data and testing on nighttime data, our Night-TTA approach showcases remarkable performance superiority over UDA methods, specifically achieving a significant \textbf{6.05}\% improvement in comparison to MS-UDA. These results highlight the efficacy and professionalism of our Night-TTA framework in addressing the challenges of domain adaptation in the context of semantic segmentation for nighttime scenarios.

{\bf The modified KP dataset.} Tab. \ref{KP Test set}  and Fig. \ref{MF-1 Result} show the quantitative and qualitative results. We can conclude that the proposed Night-TTA performs better than existing nighttime color-thermal image semantic segmentation methods. Specifically, our Night-TTA framework achieves the best segmentation performance in most categories. In addition, our proposed learning scheme for the TTA framework improves the segmentation performance of the source model (from \textbf{36.35} \% mIoU to \textbf{47.77} \% mIoU) more significantly than other TTA methods (The highest increase to \textbf{45.08}\% mIoU).



\vspace{-12pt}
\subsection{Ablation Studies and Analysis}
\noindent{\textit{{\textbf{1) Imaging Heterogeneity Refinement}}}}

{\bf \romannumeral1) Interaction Branch.}
We validate the effectiveness of the proposed interaction branch on the MF-1 dataset. The results are shown in Tab. \ref{tab:ablation studies}. During the assessment of single-modal nighttime semantic segmentation, our findings indicate that thermal imaging exhibits superior performance compared to color imaging, highlighting its heightened robustness and reliability in low-light environments. Compared with single-modal nighttime image semantic segmentation, multi-modal (color-thermal) achieves better performance. Besides, the dual path (without the interaction branch) worsens the segmentation performance (from \textbf{49.71}\% mIoU to \textbf{32.16} \% mIoU when using EEF), demonstrating the interaction branch's effectiveness.

\noindent{\bf \romannumeral2) CMSA.}
We conduct additional experiments to validate the efficacy of the CMSA module, comparing its performance in an interaction-only network and a complete network. The results, presented in Tab. \ref{tab:ablation studies}, demonstrate the significant improvements achieved by the CMSA module in both the interaction-only network (from \textbf{35.82}\% mIoU to \textbf{41.26}\% mIoU) and the triple branches networks (from \textbf{49.71}\% mIoU to \textbf{52.06}\%).

\begin{table}
\renewcommand\arraystretch{1.5}
\caption{Ablation of loss functions during Adaptation}
\vspace{-8pt}
\centering
\setlength{\tabcolsep}{2mm}
\footnotesize
\begin{tabular}{@{}ccccc>{\columncolor{gray!20}}c@{}}
\hline
Method & $\mathcal{L}^{tta}$ & $\mathcal{L}_{EN}^{tta}$ & $\mathcal{L}_{KL}^{tta}(i,EN)$ & mIoU & $\Delta$\\
\hline
Source-only & & & & 40.09 & -\\
\hline
\multirow{7}{*}{Adaptation} & \ding{51} & & & 50.07 & +9.98\\
& & \ding{51} & & 50.03 & +9.94\\
& & & \ding{51} & 50.08 & +9.99\\
& \ding{51} & \ding{51} & & 50.28 & +10.19\\
& \ding{51} &  & \ding{51} & 50.54 & +10.45\\
&  & \ding{51}  & \ding{51} & 50.83 & +10.74\\
& \ding{51} & \ding{51}  & \ding{51} & \textbf{53.16} & \textbf{+13.07}\\
\bottomrule
\end{tabular}
\vspace{-8pt}

\label{tab:Learning Scheme in Training and Testing}
\vspace{-5pt}
\end{table}

\noindent{\textit{{\textbf{2) Class Aware Refinement}}}}

{\bf \romannumeral1) EEF module.}
We compare EEF module against different methods of generating the ensemble logits (as shown in Tab. \ref{tab:ablation studies}). 
The 'Merge' approach represents taking the mean of the logits from the three branches, while 'IE' refers to methods based on image-level entropy (\eg \cite{shin2022mm}). The results demonstrate that our EEF module performs better than other strategies, with an increase of \textbf{5.64}\% (from \textbf{47.52}\% mIoU to \textbf{53.16}\% mIoU) for 'Merge' and \textbf{6.79}\%(from \textbf{46.37}\% mIoU to \textbf{53.16}\% mIoU) for 'IE' in mIoU. This highlights the superior performance of our EEF module in ensemble logits generation.

\begin{table}[t!]
\renewcommand\arraystretch{1.5}
\caption{Ablation study on effects of different architecture, CMSA module, and fusion strategy. Interaction means the Interact of color and thermal input, Ours refers to the three branches of Color, Thermal, and Interaction.}
\vspace{-8pt}
\centering

\begin{tabular}{cccccccc}
        \hline
        \multirow{2}{*}{Evaluation} & \multicolumn{3}{c}{Branch} & \multirow{2}{*}{CMF} & \multirow{2}{*}{CBF} & \multirow{2}{*}{DW} & \multirow{2}{*}{mIoU}\\\cmidrule{2-4}
        &C & I & T & \\
        \hline
        \multirow{8}{*}{CMSA} & & \ding{51} & &  &  & & 35.82\\
        && \ding{51} & & CM-FRM &  & & 37.64\\
        && \ding{51} & & CMSA &  & & 41.26\\
        &\ding{51}& \ding{51} &\ding{51} &  & EEF & & 49.71\\
        &\ding{51}& \ding{51} &\ding{51} & CM-FRM & EEF & & 50.12\\
        &\ding{51}& \ding{51} &\ding{51} & CMSA (w/o spatial) & EEF & & 51.26\\
        &\ding{51}& \ding{51} &\ding{51} & CMSA (w/o channel) & EEF & & 51.35\\
        &\ding{51}& \ding{51} &\ding{51} & CMSA & EEF & & 52.06\\
        \hline
        \multirow{4}{*}{EEF}& \ding{51}& \ding{51} &\ding{51} & CMSA  & Merge & & 47.52\\
        &\ding{51}& \ding{51} &\ding{51} & CMSA & IE &  & 46.37\\
        &\ding{51}& \ding{51} &\ding{51} & CMSA & ME & & 45.38\\
        &\ding{51}& \ding{51} &\ding{51} & CMSA & EEF & & 52.06\\
        \hline
       DW& \ding{51}& \ding{51} &\ding{51} & CMSA& EEF &\ding{51}& 53.16\\
        \bottomrule
\end{tabular}
\label{tab:ablation studies}
\vspace{-10pt}
\end{table}

{\bf \romannumeral2) Learning Scheme.}


In this experiment, the $\lambda_1$ and $\lambda_2$ are set to 1. Tab. \ref{tab:Learning Scheme in Training and Testing} shows the quantitative results. Based on our experimental data, it is evident that utilizing individual losses alone or combining any two losses leads to performance improvement in adaptation. Specifically, the three $\mathcal{L}^{tta}$, $\mathcal{L}_{EN}^{tta}$, and $\mathcal{L}_{KL}^{tta}(i,EN)$ contribute similarly during TTA, while $\mathcal{L}_{KL}^{tta}(i,EN)$ plays a slightly more important role compared with others. It should be noted that our learning scheme could significantly improve the performance of the source model (\textbf{13.07}\% mIoU). 

\noindent{\textit{{\textbf{3) Sensitivity Analysis}}}}

\noindent{\bf \romannumeral1) Batch size.}
We explore the impact of batch size on the semantic segmentation performance of different TTA methods (as shown in Tab. \ref{tab: batch size and robustness}). The results indicate that a small batch size (1 or 2) leads to degraded segmentation performance, while a larger batch size (4 or 8) results in improved performance. Tab. \ref{tab: batch size and robustness} shows that the TTA method looks very sensitive to batch size. This sensitivity can be attributed to the parameters updated by the TTA method during the test phase, primarily within the batch normalization layer. Increasing the batch size brings the testing data in a batch closer to the real data contribution during the adaptation process, thus improving the segmentation performance. The proposed method consistently performs well across different batch sizes.
It outperforms the other evaluated TTA methods in terms of mIoU, showcasing its effectiveness in semantic segmentation tasks. For example, at a batch size of 8, the proposed method achieves mIoU of 53.16, surpassing the mIoU of the other methods (ranging from 49.28 to 50.05).


\noindent{\bf \romannumeral2) Robustness to perturbations.}
We further evaluate the robustness of our methods on the MF dataset. We conduct an ablation study to evaluate the impact of different input perturbations during the test-time adaptation. Three types of perturbations are applied: image cropping, brightness adjustment, and the addition of Gaussian noise. Specifically, we crop the image at the rate of 0.2, randomly add Gaussian noise (noise range is set to 5) to the image, or just the brightness of the images to reorganize three new test sets. Tab. \ref{tab: batch size and robustness} shows the quantitative results of different TTA methods. shows the quantitative results of different TTA methods. We can conclude that our method is more robust to noises and image corruption.

\noindent{\bf \romannumeral3) Parameters updated in TTA.}
We conduct an analysis of the TTA performance by examining the impact of updating specific network layers. The ablation study aims to analyze the impact of updating specific network layers during TTA in semantic segmentation. Three scenarios are considered: updating only the encoder parameters, updating only the decoder parameters, and updating both the encoder and decoder parameters. The experiment is conducted with a batch size of 8. Tab. \ref{tab:parameters} presents the results according to updating the affine parameters in different network parts for effective TTA. When only the encoder parameters are updated during TTA, the method achieved the mIoU of 48.71. Updating only the decoder parameters result in the best performance, with a mIoU of 53.16. 

\begin{table}
\centering
\renewcommand\arraystretch{1.5}
\caption{The impact of batch size (BS) on
the top and input perturbation (bottom) on the TTA performance.}
\vspace{-8pt}
\label{tab: batch size and robustness}
\setlength{\tabcolsep}{3mm}
\footnotesize
\begin{tabular}{@{}ccccc@{}}
\hline
 & Tent\cite{wang2020tent} & MMTTA\cite{shin2022mm} & xMUDA-pl\cite{jaritz2020xmuda} & \cellcolor{gray!20}Ours\\
 \hline
BS=1 & 28.14 & 28.14 & \textbf{28.15} & \cellcolor{gray!20}\textbf{28.15}\\
BS=2 & 37.45 & 36.79 & 36.51 & \cellcolor{gray!20}\textbf{37.51}\\
BS=4 & 44.9 & 45.32 & 46.01 & \cellcolor{gray!20}\textbf{46.18}\\
BS=8 & 49.28 & 50.05 & 49.87 & \cellcolor{gray!20}\textbf{53.16}\\
 \hline
 \hline
Crop & 38.53 & 40.72 & 39.09 & \cellcolor{gray!20}\textbf{41.67}\\
Brightness & 48.07 & 49.26 & 48.25 & \cellcolor{gray!20}\textbf{51.58}\\
Noise & 47.19 & 49.20 & 48.08 & \cellcolor{gray!20}\textbf{51.37}\\
\bottomrule
\end{tabular}
\vspace{-5pt}
\end{table}

\begin{table}
\renewcommand\arraystretch{1.5}
\caption{Ablation study of  parameters updated during the TTA}
\label{tab:parameters}
\vspace{-8pt}
\centering
\setlength{\tabcolsep}{4mm}
\begin{tabular}{ccccc}
\hline
Method & Car & Person & Bike & mIoU\\  \hline
Encoder & 69.24 & 53.7 & 23.19 & 48.71\\
Decoder & \textbf{76.26} &\textbf{ 58.31} & \textbf{24.76} & \textbf{53.16}\\
Both & 70.98 & 53.95 & 22.19 & 49.04\\
\bottomrule
\end{tabular}
\end{table}

\vspace{-12pt}
\section{Discussion}

\begin{figure}[t!]
    \centering
    \includegraphics[width=0.5\textwidth]{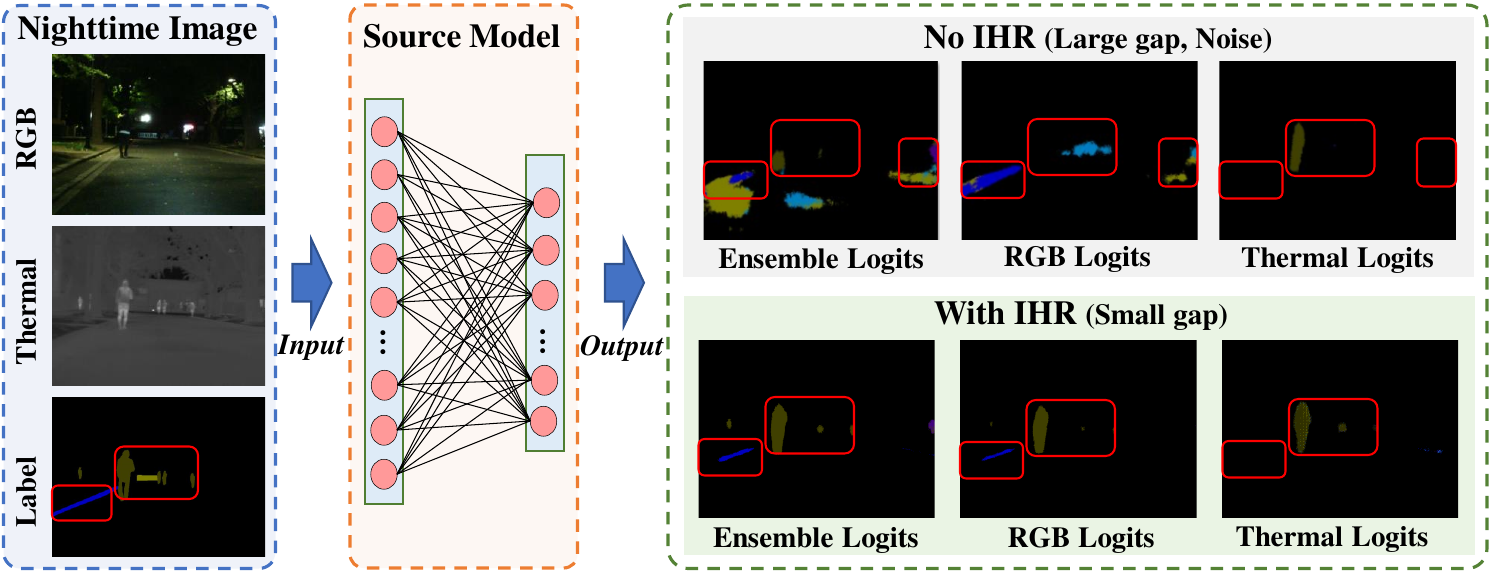}
    \vspace{-20pt}
    \caption{The efficacy of the IHR strategy becomes evident during the assessment using nighttime images. In scenarios where the IHR strategy is absent, the acquired results tend to exhibit noise. Conversely, with the incorporation of the IHR strategy, the delineation of boundaries and the semantic information within the outcomes are notably enhanced, leading to improved accuracy.}
    \label{IHR}
\end{figure}

For the IHR strategy, naively combining the individual color and thermal branches yields subpar performance due to modality gap and noise (Fig. \ref{IHR}). The proposed IHR strategy enhances prediction reliability by incorporating an interaction branch and a CMSA module. The CMSA module effectively combines cross-modal invariant features while suppressing noisy information between color and thermal modalities. Evaluating with nighttime color-thermal image pairs, we observe a performance gap between color and thermal branch logits without IHR, along with considerable noise in ensemble logits. By introducing the interaction branch and CMSA module, the discrepancy between color and thermal branch logits decreases, resulting in ensemble logits that align better with ground truth labels. This reduction in cross-modal discrepancy highlights the effectiveness of the interaction branch in mitigating the influence of image heterogeneity.

As the first TTA framework, we design three branches to generate reliable pseudo labels without considering much about the parameters and computational costs, which is typical for other cross-modal TTA methods, \eg, \cite{shin2022mm}. Future work will focus more on designing tight frameworks. Moreover, while our TTA framework is specifically designed for nighttime color-thermal semantic segmentation, there is potential for its application to address other types of multi-modality data. For instance, it can be extended to handle data combinations such as color and event data or color and depth data, opening up opportunities for broader applicability.

\begin{table}[t!]
\renewcommand\arraystretch{1.5}
\caption{Computation complexity comparisons with source-only supervised methods and TTA methods based on MF-1 dataset.}
\centering
\setlength{\tabcolsep}{1.5mm}
\footnotesize
\begin{tabular}{ccccc}
\hline
Dataset & Method  & Adapt & Time (ms) & Comput. (GFLOPs)\\
\hline
\multirow{7}{*}{MF-1} & Source-only &-& 34.2 & 291.70\\
\hline
& + LAME~\cite{boudiaf2022parameter} & TTA &40.5& 291.70\\
& + BN-Adapt~\cite{schneider2020improving} &TTA & 78.2&291.70\\
& + Tent\cite{wang2020tent} & TTA &78.1&291.70\\
& + xMUDA-pl\cite{jaritz2020xmuda} & MMTTA& 75.2& 658.62\\
& + MMTTA\cite{shin2022mm}& MMTTA & 91.8 & 974.17\\
& + Night-TTA (Ours) &MMTTA & 79.1 & 779.25\\
\bottomrule
\end{tabular}
\vspace{-10pt}

\label{tab:comparative studies_time}
\end{table}

\vspace{-12pt}
\section{Conclusion}
In this paper, we addressed two potential problems of nighttime color-thermal image semantic segmentation to reduce the cross-modal discrepancy via test time adaptation (TTA) with cross-modal ensemble distillation. We presented a novel TTA framework, dubbed \textbf{Night-TTA}, with two novel refinement strategies: imaging heterogeneity refinement (IHR) and class-aware refinement (CAR). In the experiments, both strategies were shown effective in achieving credible performance. The experimental results also proved the benefits of our learning scheme. Moreover, for nighttime color-thermal semantic segmentation, Night-TTA outperformed the existing methods by a considerable margin. 

\vspace{-12pt}

\section{Acknowledgment}
This work was supported by the National Natural Science Foundation of China (NSF) under Grant No. NSFC22FYT45 and 2023 Guangzhou City- Enterprise Joint Funding Scheme under Grant No. SL2022A03J01278.

{
\bibliographystyle{IEEEtran}
\bibliography{ref}
}

\begin{IEEEbiography}[{\includegraphics[width=0.8in,height=1.0in,clip,keepaspectratio]{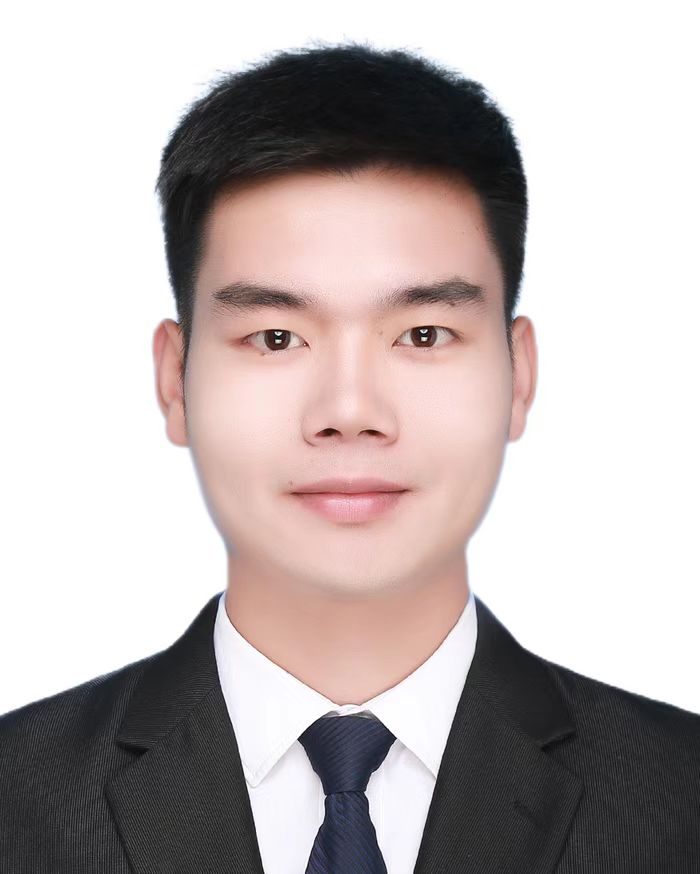}}] {Yexin Liu} is a Mphil. student in the Visual Learning and Intelligent Systems Lab,  Artificial Intelligence
Thrust, The Hong Kong University of Science and Technology,  Guangzhou (HKUST-GZ).
His research interests include infrared- and event-based vision, and unsupervised domain adaptation.
\end{IEEEbiography}

\begin{IEEEbiography}[{\includegraphics[width=0.8in,height=1.0in,clip,keepaspectratio]{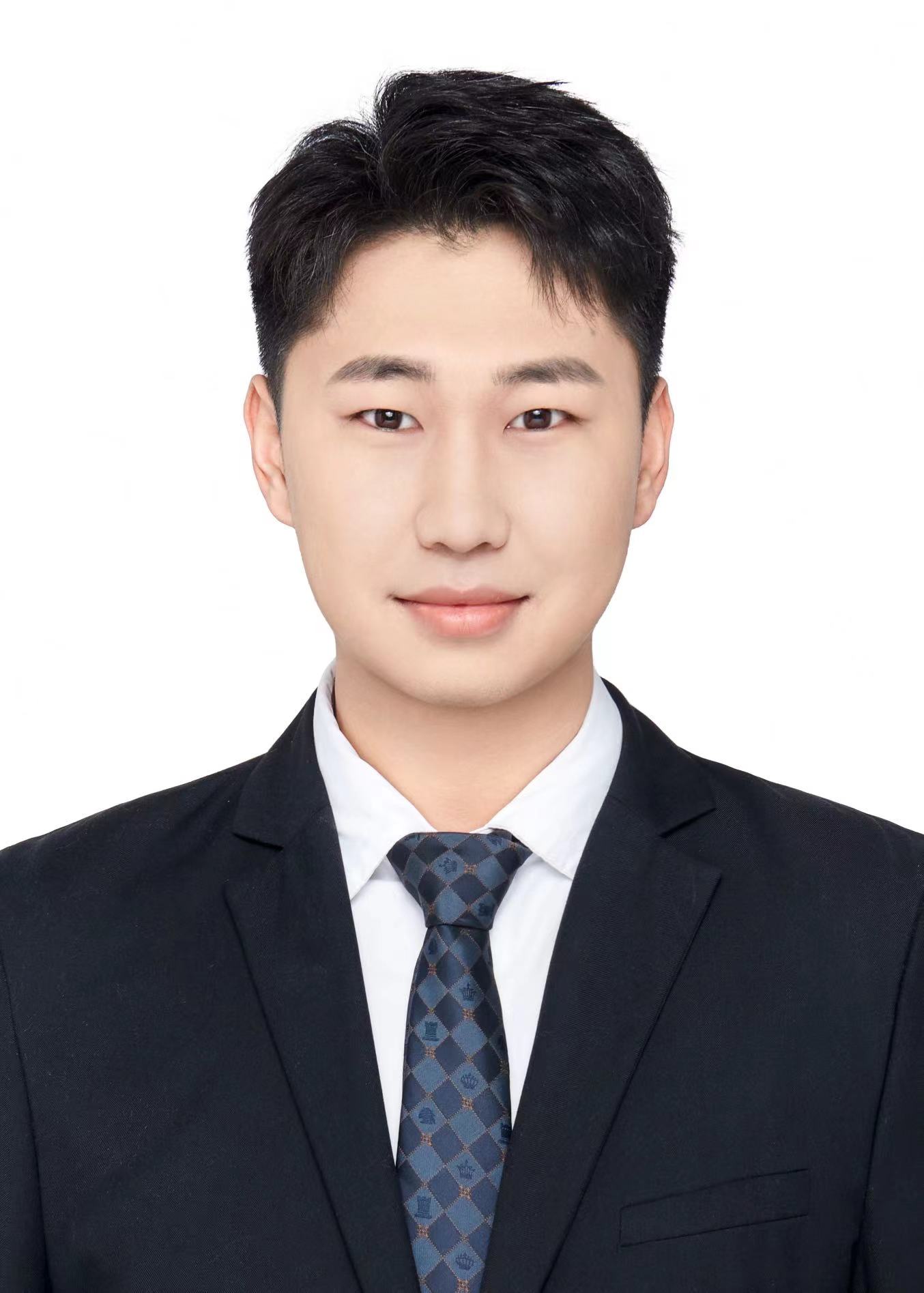}}] {Weiming Zhang}
is a research assistant in the Visual Learning and Intelligent Systems Lab,  Artificial Intelligence
Thrust, The Hong Kong University of Science and Technology,  Guangzhou (HKUST-GZ).
His research interests include event-based vision, Deep Learning, \etc.
\end{IEEEbiography}

\begin{IEEEbiography}[{\includegraphics[width=0.8in,height=1.0in,clip,keepaspectratio]{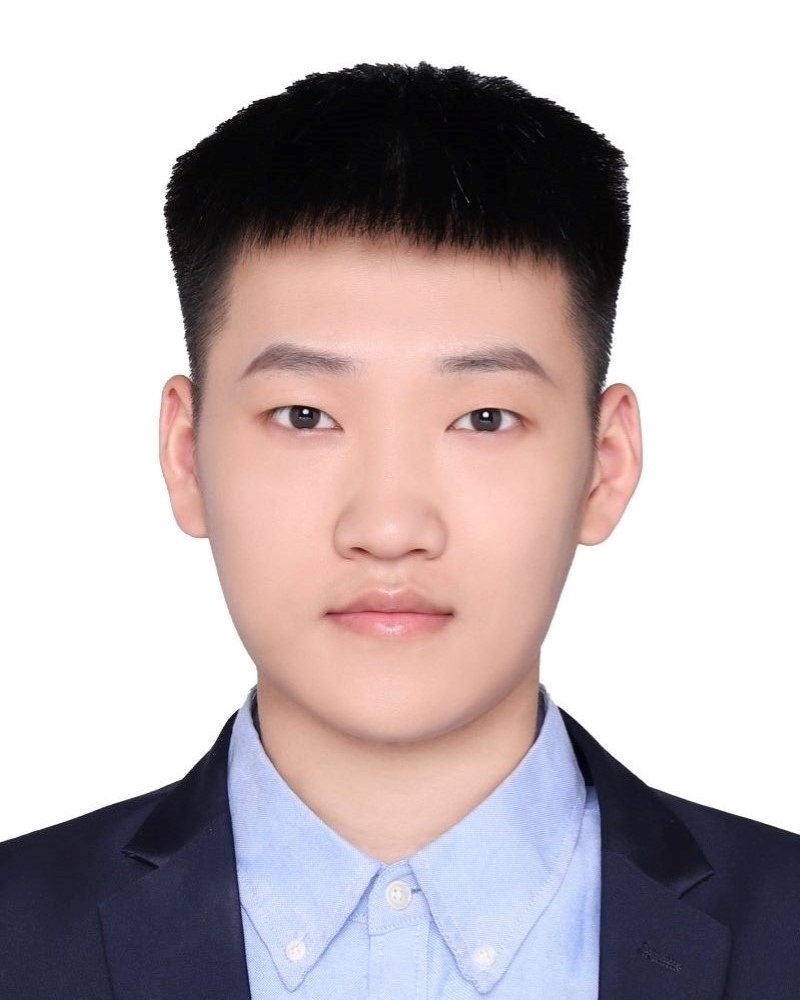}}] {Guoyang ZHAO} is a Mphil. student in the Intelligent Autonomous Driving Center, Thrust of Robotics and Autonomous Systems, The Hong Kong University of Science and Technology,  Guangzhou (HKUST-GZ). His research interests include vision-based perception system and Deep learning.
\end{IEEEbiography}

\begin{IEEEbiography}[{\includegraphics[width=0.8in,height=1.0in,clip,keepaspectratio]{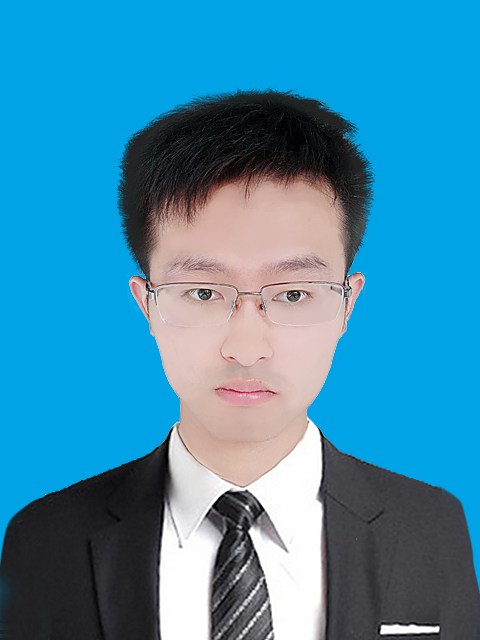}}] 
{Jinjing Zhu} is a Ph.D. student in the Visual Learning and Intelligent Systems Lab, Artificial Intelligence Thrust, The Hong Kong University of Science and Technology,  Guangzhou (HKUST-GZ). His research interests include CV (image classification, person re-identification, action recognition, etc.), DL (especially transfer learning, knowledge distillation, multi-task learning, semi-/self-unsupervised learning, etc.), omnidirectional vision, and event-based vision.
\vspace{-10pt}
\end{IEEEbiography}

\begin{IEEEbiography}[{\includegraphics[width=0.8in,height=1.0in,clip,keepaspectratio]{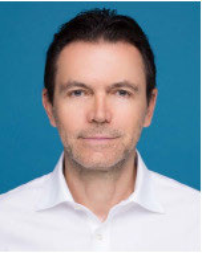}}] 
{Athanasios V. Vasilakos} is with the Center for AI Research (CAIR), University of Agder(UiA), Grimstad, Norway. He served or is serving as an Editor for many technical journals, such as the IEEE TRANSACTIONS ON AI, IEEE TRANSACTIONS ON NETWORK AND SERVICE MANAGEMENT; IEEE TRANSACTIONS ON CLOUD COMPUTING, IEEE TRANSACTIONS ON INFORMATION FORENSICS AND SECURITY, IEEE TRANSACTIONS ON CYBERNETICS; IEEE TRANSACTIONS ON NANOBIOSCIENCE; IEEE TRANSACTIONS ON INFORMATION TECHNOLOGY IN BIOMEDICINE; ACM Transactions on Autonomous and Adaptive Systems; the IEEE JOURNAL ON SELECTED AREAS IN COM-MUNICATIONS . He is WoS highly cited researcher(HC).
\end{IEEEbiography}

\begin{IEEEbiography}[{\includegraphics[width=0.8in,height=1.0in,clip,keepaspectratio]{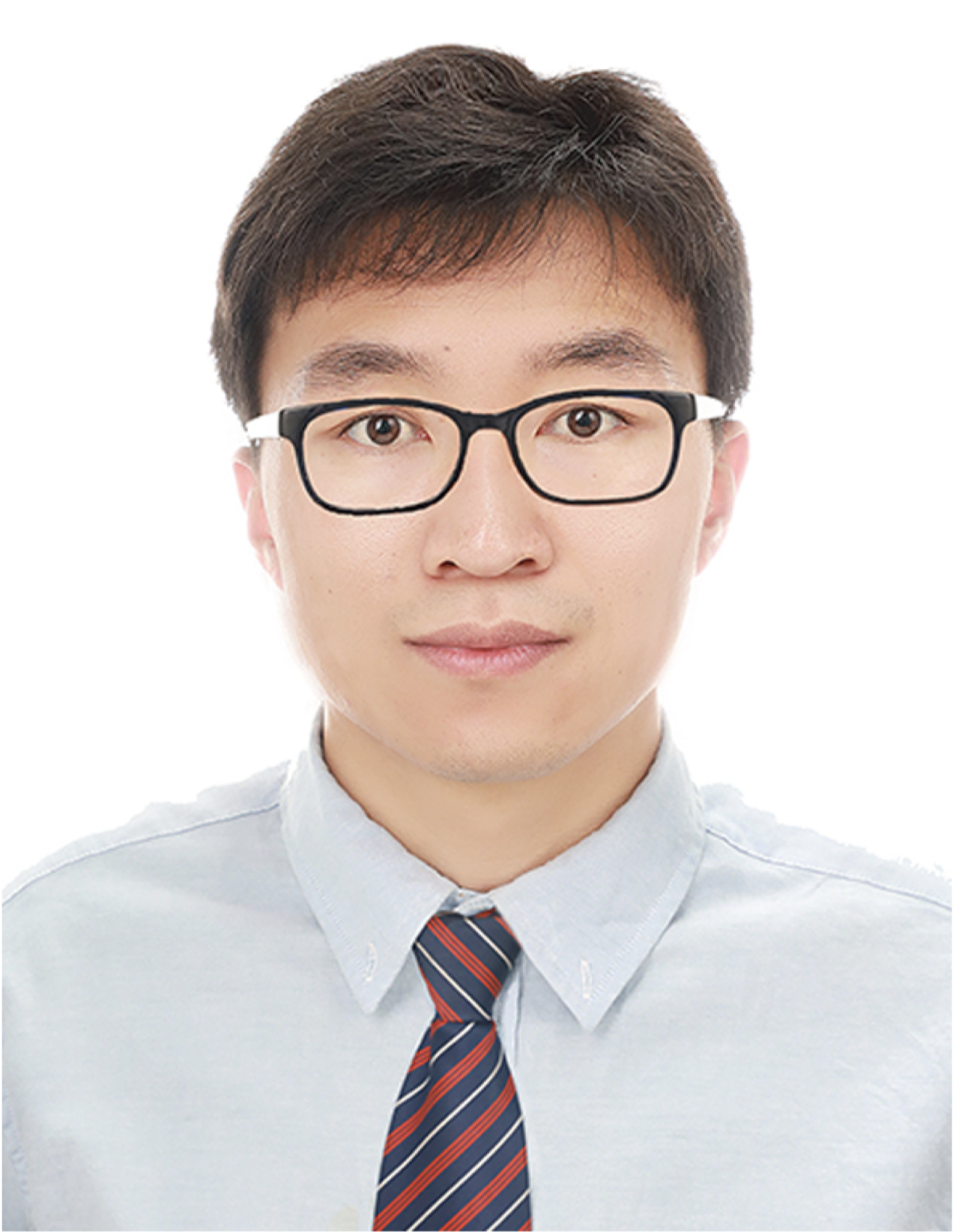}}] 
{Lin Wang} (IEEE Member) is an assistant professor in the AI Thrust, HKUST-GZ, HKUST FYTRI, and an affiliate assistant professor in the Dept. of CSE, HKUST. He did his Postdoc at the Korea Advanced Institute of Science and Technology (KAIST). He got his Ph.D. (with honors) and M.S. from KAIST, Korea. He had rich cross-disciplinary research experience, covering mechanical, industrial, and computer engineering. His research interests lie in computer and robotic vision, machine learning, intelligent systems (XR, vision for HCI), etc.  
\end{IEEEbiography}

\end{document}